\documentclass[journal]{IEEEtran}
\usepackage{amsmath,amsfonts}
\usepackage{mathtools}
\usepackage{algorithmic}
\usepackage{algorithm}
\usepackage{array}
\usepackage[caption=false,font=normalsize,labelfont=sf,textfont=sf]{subfig}
\usepackage{textcomp}
\usepackage{stfloats}
\usepackage{url}
\usepackage{verbatim} 
\usepackage{graphicx}
\usepackage{cite}
\usepackage{booktabs}
\usepackage{upgreek}
\usepackage{multirow}
\usepackage{bm} 
\usepackage[para]{threeparttable}
\usepackage{siunitx}
\usepackage{lipsum}
\begin{document}

\title{Electromagnets Under the Table: an Unobtrusive Magnetic Navigation System for Microsurgery}

\author{ Adam Schonewille, Changyan He, Cameron Forbrigger, Nancy Wu, \\ James Drake, Thomas Looi, Eric Diller

\thanks{A. Schonewille, C. He, N. Wu, and E. Diller are with the Department of Mechanical and Industrial Engineering at the University of Toronto, Canada.}
\thanks{C. He, N.Wu, J. Drake, and T. Looi are with the PCIGITI center at the SickKids Hospital, Canada.}
\thanks{C. Forbrigger was with the Department of Mechanical and Industrial Engineering at the University of Toronto, Canada. He is now with the Department of Health Sciences and Technology at ETH Zürich, Switzerland.}
\thanks{E. Diller is also with the Robotics Institute and Institute of Biomedical Engineering at the University of Toronto, Canada.}
\thanks{(ediller@mie.utoronto.ca)}}

\maketitle

\begin{abstract}
Miniature magnetic tools have the potential to enable minimally invasive surgical techniques to be applied to space-restricted surgical procedures in areas such as neurosurgery. However, typical magnetic navigation systems, which create the magnetic fields to drive such tools, either cannot generate large enough fields, or surround the patient in a way that obstructs surgeon access to the patient. This paper introduces the design of a magnetic navigation system with eight electromagnets arranged completely under the operating table, to endow the system with maximal workspace accessibility, which allows the patient to lie down on the top surface of the system without any constraints. 
The found optimal geometric layout of the electromagnets maximizes the field strength and uniformity over a reasonable neurosurgical operating volume. The system can generate non-uniform magnetic fields up to 38 mT along the $x$ and $y$ axes and \mbox{47 mT} along the $z$ axis at a working distance of 120 mm away from the actuation system workbench, deep enough to deploy magnetic microsurgical tools in the brain. 
The forces which can be exerted on millimeter-scale magnets used in prototype neurosurgical tools are validated experimentally. Due to its large workspace, this system could be used to control milli-robots in a variety of surgical applications.
\end{abstract}


\section{Introduction}

\IEEEPARstart{M}{edical} microrobots have been explored in the past two decades for their potential to operate wirelessly within small areas of the body where conventional tools struggle to reach. This ability to control objects at very small scales has applications in a variety of fields. In particular, these newly developed devices show potential to push the boundaries of medicine in applications such as disease diagnostics, biopsies, targeted drug delivery, and minimally invasive microsurgeries \cite{Diller2015}. Challenges in fabrication, high friction and energy storage lead to micro/milli robots which are relatively simple. Magnetic fields are often used to drive such microrobots for medical use, as low-frequency magnetic fields can safely penetrate deep into the human body without distortion or attenuation by tissue, without causing heat generation in tissue. These magnetic fields are used to apply forces and torques to small magnets embedded inside a microrobot, which serves as an untethered end effector of the system. 

Magnetic fields for actuating magnetic devices are often generated in labs using a Helmholtz coil system \cite{Ref-9}, which surrounds the workspace and can generate a uniform magnetic field. Such electromagnetic coil arrangements typically generate only low magnetic field magnitudes. There are several solutions to increase the magnetic field that all involve increasing the strength of the magnetic actuator. Large permanent magnets can be employed to generate strong fields with relatively simple supporting structures, but they present a safety hazard as their magnetic field can not be turned off in case of emergency. For electromagnetic systems, adding a soft-magnetic core material and operating at a higher current can improve the output magnetic field flux. These changes can be seen in the clinical scale actuation system developed by Philips GmbH \cite{ref-38} as well as in the OctoMag system \cite{ref-17}. These systems, however, suffer from poor access to the workspace. Dynamic (movable) systems such as BigMag \cite{ref-44} or the Stereotaxis Niobe System \cite{ref-4} have large workspaces thanks to their electromagnets' mobility. However, they are typically cumbersome and need complex motorized actuation, reducing their adaptability to operating rooms.

\label{Section-WorkspaceAccessibility}
For use in a surgical setting, an open workspace is highly desirable. Here we introduce a metric to define workspace accessibility for electromagnetic navigation systems. We define workspace accessibility as the physical space available and not occupied by coils, as measured by solid angles, in a manner similar to Pourkand et al. \cite{ref-37}. We define the maximum workspace accessibility as the single largest solid angle of open space for the whole system. From the microrobot location, the solid angle $\Phi$ is defined by the angle of a conical view projecting out of the system through the largest opening between actuators. 
 
The maximum workspace accessibility for a number of relevant magnetic actuation system by this definition is shown in \mbox{Table \ref{Table-WorkspaceCompare}}. Only systems that have 8 magnetic actuators are surveyed, since this is the minimum number of actuators needed to achieve full control over all 8 magnetic field degrees of freedom (DOF) to allow full control of sophisticated magnetic robotic end effectors \cite{Ref-36}. Thus, systems such as \mbox{NavionMag \cite{CardioMag-NavionMag}} with only three magnets are not included.

\begin{table}[bt]
\begin{threeparttable}[b]
\centering
\caption{Ranked workspace accessibility of eight-coil magnetic actuation systems}
\begin{tabular}{c|c}
\hline
Magnetic Actuation System        & Workspace Accessibility $\Phi$ \\ \hline
\textbf{This Work}                        & $222 ^{\circ}$                 \\
Rahmer et al. \cite{ref-38}           & $<90 ^{\circ}$\tnote{a}        \\
OctoMag \cite{ref-17}               & $120 ^{\circ}$                 \\
Open-asymmetric OctoMag \cite{ref-37} & $100 ^{\circ}$                 \\
CardioMag \cite{CardioMag-NavionMag}    & $<90 ^{\circ}$                 \\   
MiniMag \cite{minimag}               & $220 ^{\circ}$                 \\
Salmanipour et al. \cite{Ref-40}     & $90 ^{\circ}$                 \\
BatMag \cite{Ref-32}                 & $<90 ^{\circ}$\tnote{b}       \\
Square Antiprism \cite{Ref-14}      & $48 ^{\circ}$                 \\
Square Prism \cite{Ref-24}        & $40 ^{\circ}$                 \\ 
Hwang et al. \cite{HongsooChoi}    & $120 ^{\circ}$                 \\\hline
\end{tabular}
\begin{tablenotes}[normal,flushleft]
       \item [a] We assume that the cylinder the patient fits into is at most the same length as diameter although it looks longer in the figures.
       \item [b] Adequate information is not given about the orientations of the magnets in this system, but the system resembles a square prism layout with more actuators so the accessibility is at most twice the accessibility of a square prism design.
 \end{tablenotes}
\label{Table-WorkspaceCompare}
\end{threeparttable}
\end{table}

It is clear from Table \ref{Table-WorkspaceCompare} that the MiniMag system and the OctoMag system are the systems that have the highest workspace accessibility. Since all the surveyed electromagnetic systems shown here either completely surround the operating space, or partially surround the operating space in a hemispherical manner, almost all of the workspace accessibility values are less than $180^\circ$. Furthermore, lower accessibility values also reduce the size of object which can fit inside the workspace, meaning that a human patient may not fit. A magnetic actuation system with all the actuators below a plane would have much higher accessibility than a system whose actuators partially surround the operating space in a concave manner. This will be beneficial in a clinical setting in terms of flexibility in patient positioning and surgeon access to the patient from any angle above the plane. Surgeon access is crucial if, for example, something goes wrong during a procedure and the surgeon needs to switch to manual tools. 

Table-like magnetic actuation systems exist in the literature, but they are not designed for the purpose of manipulating the magnetic field over large distances. MagTable \cite{Ref-57} consists of many magnetic actuators that generate localized fields to translate a microrobot along a plane instead of manipulating it in one place. Its actuators are not strong enough to generate fields over large distances for medical applications. A highly reconfigurable magnetic actuation system that fits the table-like design is the OmniMagnet \cite{Ref-35}. This actuation system consists of tri-nested electromagnets with a spherical core. This system does not have enough magnets for full control over the field components, but additional duplicates of the system can be used in tandem to effectively achieve the desired control. The downsides to this device are that the fields of 3 mT at 120 mm are relatively weak. The compact design would make thermal management for higher current densities difficult. Systems like the DeltaMag \cite{ref-53}, the ARMM system \cite{ref-45}, and the system developed in \cite{DualPM-TRO-Valdastri} do not have a readily-defined workspace because the actuators are mobile. Such systems have variable accessibility which one could state as either very high (any position in space could be free of the actuation system) or zero (no position is always free of the actuation system). These systems allow for a lot of freedom in where the magnetic field can be applied over a large volume if the robot can reach it. The downsides are that these actuation systems do not consist of many magnetic actuators which limits the control over the microrobot, and that the moving components are operating in the vicinity of the surgical team in the operating room.

In this paper, we aim to demonstrate that a magnetic actuation system can be designed to maximize workspace accessibility without seriously compromising on field strength by utilizing non-uniform magnetic fields for control, which is supported through experimentation and the development of a new magnetic actuation system. We show how a magnetic actuation system can be designed without optimizing for isotropy in its magnetic field generation as is typically done in the literature. Additionally we demonstrate how this system can be used to increase the maximum magnitude of the magnetic field by a factor of 2-3 by allowing non-uniform magnetic fields as opposed to requiring uniform magnetic fields for controlling the microrobot. Due to the control methodology used, these design choices are beneficial to the actuation if the microrobot is tethered and solely actuated by magnetic torques, and benefits further if it is not necessary to have equal strength actuation in all directions in space.

The novel work of this paper contributes to the field of magnetic actuation of medical microrobots by:
\begin{itemize}
    \item Developing a magnetic actuation system with the largest unobstructed workspace and full magnetic rank by using a new design methodology that does not aim to maximize field isotropy.
    \item Demonstrating the ability to use non-uniform fields over uniform fields to increase the maximum magnetic field which can be generated when controlling a tethered microrobot.
    \item Providing an analysis on the influence of field non-uniformity on a tethered microgripper. 
\end{itemize}

\section{Magnetic Methods Review}
This section provides a review of the mathematical background for controlling the magnetic fields produced by various magnetic actuators \cite{Abbott2020}. In this paper, any matrix with \mbox{a $^\dag$} represents the pseudoinverse of the matrix. The gradient operator represents partial derivatives in each of the three basis directions of a given frame, which takes the form $\nabla = [\frac{\partial}{\partial x}, \frac{\partial}{\partial y}, \frac{\partial}{\partial z}] ^\text{T}$.

Electromagnets only generate a magnetic field when energized with an applied current. Since the field is produced by applying current, the strength and direction of the electromagnet’s magnetization can be altered by changing the magnitude and direction of the current flow, respectively. When a \mbox{current $i$} is passed through a conductor, such as a copper wire, a magnetic field \textbf{B} at any position ${\bf P}$ is produced. The field will induce torques in magnets onboard the microrobot, while spatial gradients of this field induces forces. 

The field at every point in space can be assumed as linear with respect to the current (for air-core electromagnets). An electromagnet's field distribution induced by a current $i$ can be approximated by a dipole model, at distances slightly removed from the magnet. In a magnetic actuation system consisting of several electromagnets, the magnetic field at any position in the workspace is the linear superposition of all dipole field contributions as
\begin{equation}
    \bf{B}(\bf{P}) = \sum_{i=1}^{n} (\frac{\mu_0}{4\pi||\bf{r}_i||^3}(3\hat{\bf{r}}_i\hat{\bf{r}}_i^{\text{T}}-\mathbb{I}_3))\widetilde{\bf{m_i}},
\end{equation}
where $\bf{P}_\text{m}$ is the electromagnet position, $\bf{r}_i=\bf{P}-\bf{P}_\text{m}$, and $\widetilde{\bf{m_i}}$ is the coil dipole magnetic moment. When the magnetic field is solved for at one point for a static geometry, the magnetic field is straightforward to calculate for all applied currents by simple scaling. This linear dependence based off calibrations can be simply written for a single electromagnet as 
\begin{equation}
    {\bf B}({\bf P}) = \widetilde{\bf B} i,
    \label{Eq-Blinear}
\end{equation}
which can be extended to multiple electromagnets as 
\begin{equation}
    {\bf B}({\bf P}) = \sum_{j=1}^{n} \tilde{\bf B}_j i_j = \mathcal{B}({\bf P}) {\bf I}.
    \label{Eq-Bcalibrate}
\end{equation}
A similar method can be followed to formulate the magnetic field spatial gradients in terms of the input current(s) as
\begin{equation}
    {\bf G}({\bf P}) = \mathcal{G}({\bf P}) {\bf I}.
    \label{Eq-Gcalibrate}
\end{equation}
These cases have only concerned air-cored electromagnets, where it is valid to assume that the fields contributions of each electromagnetic actuator linearly superimpose.
To increase the magnetic flux produced by an electromagnet, it is advantageous to insert a ferromagnetic core into the design. The assumption that the magnetic field components $ \bf  B(P)$ and $\bf G(P)$ will consist of the linear superposition of all individual electromagnetic field contributions still holds true if the core material behaves as an ideal soft magnet (has low coercivity), and the core does not reach its saturation magnetization for the current densities applied [17].

For electromagnets with a linear relationship between coil current and outputs, the problem can be written as
\begin{equation}
    \left[
\begin{array}{c}
{\bf B(P)} \\
{\bf G(P)}
\end{array} \right]= \left[
\begin{array}{c}
\mathcal{B}({\bf P}) \\
\mathcal{G}({\bf P})
\end{array}  
\right] \textbf{I} = \mathcal{U}({\bf P}) \textbf{I}, 
\label{Eq-ControlMatrix-BG}
\end{equation}
where the currents necessary for producing a desired field can be determined by inverting $\mathcal{U}({\bf P}) $
\begin{equation}
 \bf I = \mathcal{U}({\bf P})^{-1}\left[
\begin{array}{c}
{\bf B_{des}(P)} \\
{\bf G_{des}(P)}
\end{array} \right].
\label{Eq-Uni-B}
\end{equation}
In this manner, with 8 well-conditioned magnetic actuators, a desired uniform magnetic field can be generated by setting $\bf G_{des}(P) = 0$. Using \eqref{Eq-Uni-B} in this way is referred to as a \textbf{uniform control methodology} throughout the paper. If the number of electromagnets is not 8 then the control matrix $\mathcal{U}({\bf P})$ is not square, and thus not invertible. In this case, the pseudo-inverse of the control matrix $\mathcal{U}({\bf P})^{\dag}$ can be used to calculate the command current.

Even if the number of actuators allows $\mathcal{U}({\bf P})$  to be square and invertible, if we only care about controlling the magnitude of the magnetic field and do not care about the gradient components, the pseudo-inverse can be used to determine the best inputs to achieve this field.
\begin{equation}
    \bf I = \mathcal{B}({\bf P})^{\dag} \bf B_{des}(P).
    \label{Eq-nonUni-B}
\end{equation}
Throughout the rest of the paper, this approach to controlling the magnetic fields will be referred to as the \textbf{non-uniform control methodology} as the gradients can take on any value that would maximize the desired fields.

A magnetic object with magnetic moment \mbox{${ \textbf{m}} = [ m_x, m_y, m_z ]^\text{T}$} experiences a force ${\bf f} $
and torque ${\bm \uptau} $
when a magnetic field ${\bf B}$ is applied as
\begin{equation}
    {\bf f} = \nabla (\bf B \cdot m), \text{ and}
    \label{Eq-MagForce}
\end{equation}
\begin{equation}
    \bm \uptau = \bf m \times \bf B.
    \label{Eq-MagTorque}
\end{equation}

\section{System Geometric Design}
\subsection{Design Objectives}
\begin{figure}[bt]
  \centering
  \includegraphics[width=0.4\textwidth]{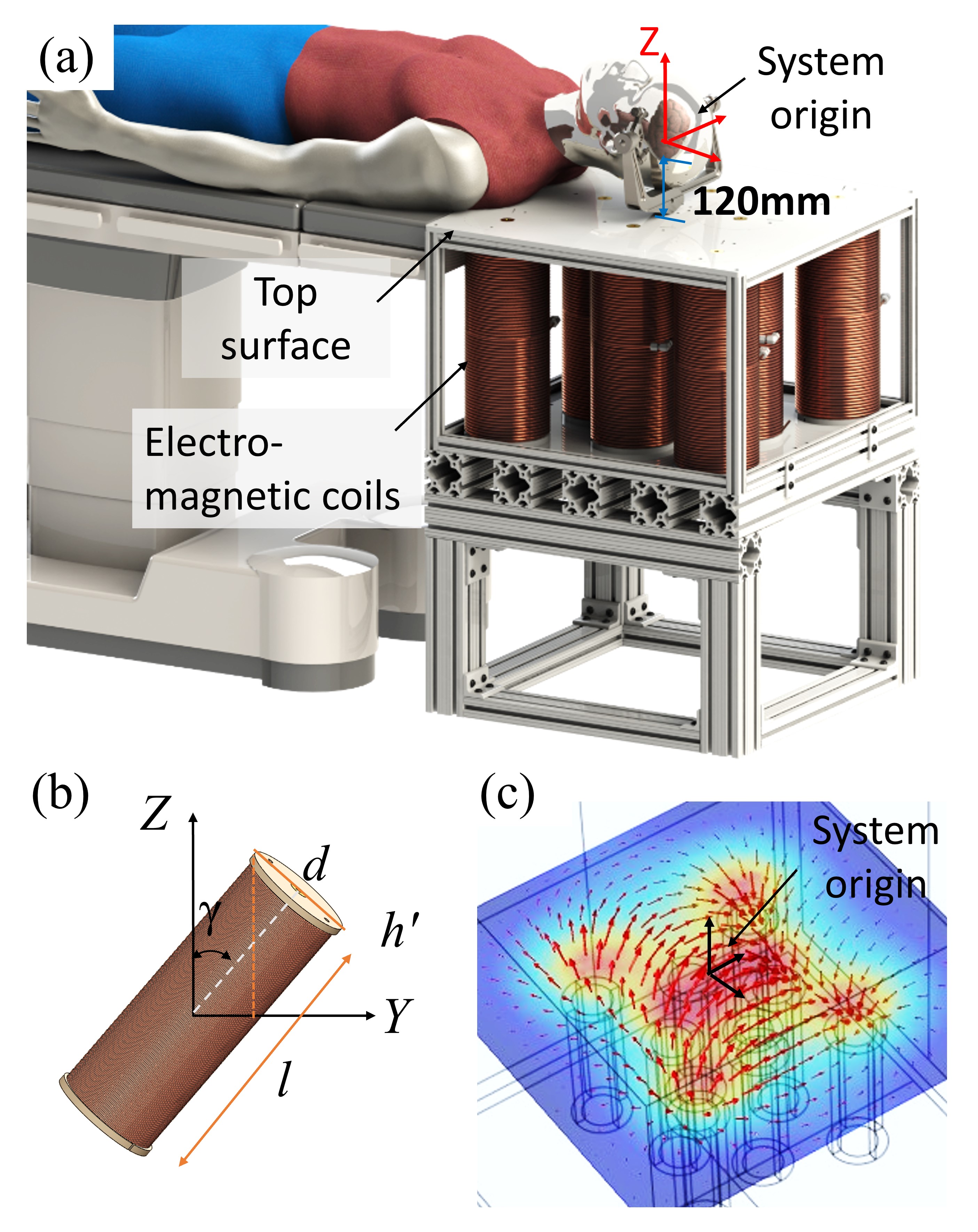}
  \caption{(a) Magnetic actuation system in its target application setting. (b) Diagram for deriving the height relationship to electromagnet orientation. The electromagnet's frame is relative to the center of the magnetic actuator. (c) Simulated magnetic field streamlines showing that the field is controlled to be unidirectional at the system origin.}
  \label{Fig-SystemConcept}
\end{figure}

A concept rendering of the coil system design is shown in Fig. \ref{Fig-SystemConcept}(a). The neurosurgeons we consulted with requested that the magnetic actuation system not surround the workspace or obstruct the view of the patient. This requirement was motivated by patient safety: when an operation is underway, the surgeons may easily step in and take manual control of the surgery if anything were to go wrong. In order to be as unobstructed to the neurosurgeon as possible, the design should avoid surrounding the patient with actuators. To maximize usable space, we chose to locate the magnets in the unused dead space that the surgical table occupies below the patient. This would allow the surgical environment to remain virtually unchanged from its current form. This design shows the most accessible workspace of any published electromagnetic navigation system with stationary electromagnets showing full control over all magnetic field components.

For the target application of neurosurgery, the workspace of the system must be large enough to fit a human head. We approximated this requirement as a spherical volume with a \mbox{240 mm} diameter based on the maximum adult head width of 239 mm reported in a study of 105 adult males in the United States \cite{Lee2006}. We assume the working distance to be the center of this sphere, or 120 mm from the surface of the table. 

\subsection{Electromagnet Geometry}
We conducted numerical simulations in MATLAB and ANSYS to determine the magnetic field output at a working distance of 120 mm for different electromagnet arrangements, where we also considered the system mass. 

For the numerical simulations in MATLAB, we set the aspect ratio (the ratio of length to the core radius) of each electromagnet to $Q \ge 8$, as this choice of aspect ratio allows us to use an infinite solenoid model to predict the magnetic field at the center of the electromagnet core with a relative error of less than $6.13\%$ \cite{ref-18}. Furthermore, for a \textsc{Vacoflux} 50 alloy core, it was shown through FE simulation that electromagnets with a smaller aspect ratio require a higher current density to saturate. 
For every size of electromagnet studied, the maximum current density for a constant input power of 1.8 kW was calculated and used to simulate the max magnetic field for a constant amount of available electrical power in our laboratory. We considered copper wire of thickness from 6 AWG to 28 AWG.

The electromagnet size which creates the largest field given these constraints is a cylinder with the core radius of 45 mm and the length of 360 mm. This electromagnet geometry with a \textsc{Vacoflux} 50 alloy core is able to produce an axial magnetic field of 63.8 mT at a distance of 120 mm when a current density of 6 A/mm$^2$ is applied. It also has an approximate mass of 38.7 kg. This electromagnet is used in the subsequent layout optimization process.

\section{Electromagnet Layout Optimization}
This section outlines the process used to design an electromagnetic actuation system that is optimized to maximize the non-uniform magnetic fields that can be generated, without consideration of the isotropy of the system. Most electromagnetic actuation systems from the literature were designed with field isotropy as an optimization priority, but our approach abandons this notion in favour of maximizing magnetic field output to maximize the actuation strength of our microsurgical tools. For all of the following simulations, the electromagnet will be modeled as a single-point dipole moment located at the centroid of the electromagnet core. 

\subsection{Objective Function and Gradient Descent}
The electromagnetic actuation system will have 8 electromagnets all subject to the same geometric constraints. Firstly, the electromagnets should all be located below the plane that defines the surface of the operating table at height $z = -0.120$ m (the center of the workspace is at the origin). The actuators should be located as close to the table surface as possible to minimize their distances from the workspace center, which will maximize their magnetic field contributions. The second constraint applied requires that all electromagnets be adequately spaced apart to be able to realize the design given requirements for sturdy mounting. These constraints will be implemented through penalty functions in the optimization algorithm.

Each electromagnet is modeled as a cylinder with 5 free location parameters expressed in spherical coordinates: 3 position parameters $x, y, z$ and 2 orientation parameters $\beta,\gamma$. For $n=8$ actuators there are thus $5n=40$ free parameters over which to optimize. These design parameters are loaded into a single vector of length 40:
\begin{equation}
\begin{split}
        \widetilde{\bf x}  =  [ & x_1, \cdots, x_n, \quad y_1, \cdots, y_n, \quad z_1, \cdots, z_n,\\ & \beta_1,  \cdots, \beta_n, \quad \gamma_1, \cdots, \gamma_n].
\end{split}
\end{equation}

For the optimization algorithm, we want to maximize the magnetic field that can be generated without enforcing the produced field to be isotropic (equal field generation capability in all directions). This is because we hypothesize that with all the electromagnets on one side of the system, the magnetic field will naturally favor the direction in which the electromagnets point, resulting in an imbalance in maximum magnetic field for different principle axes. We have noted that the actuation of many magnetic tools actually requires strong fields in only certain directions. An optimization algorithm was run to optimize two different control scenarios (uniform control and non-uniform control) with this imbalance in mind. 

The associated functions are a set of penalty and optimization functions that make up the basis of the objective function. We divided the associated functions into 3 sets of governing functions referred to as the magnetic field optimization functions $\widetilde{\bf M}(\widetilde{\bf x})$, the electromagnet height penalty functions $\widetilde{\bf H}(\widetilde{\bf x})$, and the electromagnet proximity penalty functions $\widetilde{\bf P}(\widetilde{\bf x})$. The magnetic field optimization functions $\widetilde{\bf M}(\widetilde{\bf x})$ are responsible for maximizing the magnetic fields produced. The electromagnet height penalty functions $\widetilde{\bf H}(\widetilde{\bf x})$ attempt to enforce the height constraint as a penalty which returns a minimum when the electromagnet is in contact with the upper bounding plane.
The proximity penalty functions $\widetilde{\bf P}(\widetilde{\bf x})$ keep each electromagnet away from its neighbours and avoid overlap in space. All of the associated functions are described in detail in the \textit{Supplementary File}.
These optimization functions are combined into one vector as
\begin{equation}
    \widetilde{\bf G}(\widetilde{\bf x}) = \left[
\begin{array}{c}
\widetilde{\bf M}(\widetilde{\bf x})\\
\widetilde{\bf H}(\widetilde{\bf x})\\
\widetilde{\bf P}(\widetilde{\bf x})
\end{array}
\right] =
\left[
\begin{array}{c}
\widetilde{G}_1(\widetilde{\bf x})\\
\widetilde{G}_2(\widetilde{\bf x})\\
\vdots \\
\widetilde{G}_N(\widetilde{\bf x})
\end{array}
\right],
\end{equation}
where the total number of functions is $N = 3 + \frac{n}{2}(n + 1)$ for $n$ electromagnets. For the $n=8$ electromagnets we are simulating, we will have 39 associated functions to match the 40 optimization parameters. These associated functions will be minimized by the gradient descent algorithm. For metrics that we wish to maximize, the inverse of the quantity will make up the associated function.

The set of associated functions $\widetilde{\textbf{G}}(\widetilde{ \textbf{x}})$ make up the objective function $\widetilde{F}(\widetilde{\textbf{x}})$. The objective
function is chosen as one half of the sum of squares of $\widetilde{\textbf{G}}(\widetilde{\textbf{x}})$ as
\begin{equation}
    \widetilde{F}(\widetilde{\textbf{x}}) = \frac{1}{2}\widetilde{\textbf{G}}(\widetilde{\textbf{x}})^\text{T} \widetilde{\textbf{G}}(\widetilde{\textbf{x}}).
\end{equation}

A gradient descent method with randomly chosen input parameters $\widetilde x_0$ is used to minimize this objective function, where the next set of parameters is chosen as
\begin{equation}
    \widetilde x_{j+1} = \widetilde x_{j} - \eta_0\Delta\widetilde{F}(\widetilde{\textbf{x}}_j) \\
   = \widetilde x_j - \eta_0 \textbf{J}_G(\widetilde{\textbf{x}}_j)^{\text{T}}\widetilde{\textbf{G}}(\widetilde{\textbf{x}}_j).
\end{equation}
A step size $\eta_0$ is set arbitrarily small to 0.001 which ensures convergence at a reasonable pace. Additionally, the gradient of the objective function can be rewritten as $\textbf{J}_G(\widetilde x_j)^{\text{T}}\widetilde {\textbf{G}}(\widetilde{\textbf{x}}_j)$ where $\textbf{J}_G(\widetilde{\textbf{x}})$ is the Jacobian matrix of the associated function $\widetilde{\textbf{G}}(\widetilde{\textbf{x}})$ which can be determined numerically due to the complexity of the formulas with a $\Delta \widetilde{\textbf{x}}$ value of 0.0001 for finding the numerical derivative with respect to any of the optimization parameters. The optimization stops at the condition of the objective function not descending significantly \mbox{($||\widetilde{F}(\widetilde{\textbf{x}})_{j+1}-\widetilde{F}(\widetilde{\textbf{x}})_j||<\epsilon$}, with $\epsilon$ set as $1e^{-9}$).

\subsection{System Optimization Results}
\begin{figure}[tb]
  \centering
  \includegraphics[width=0.48\textwidth]{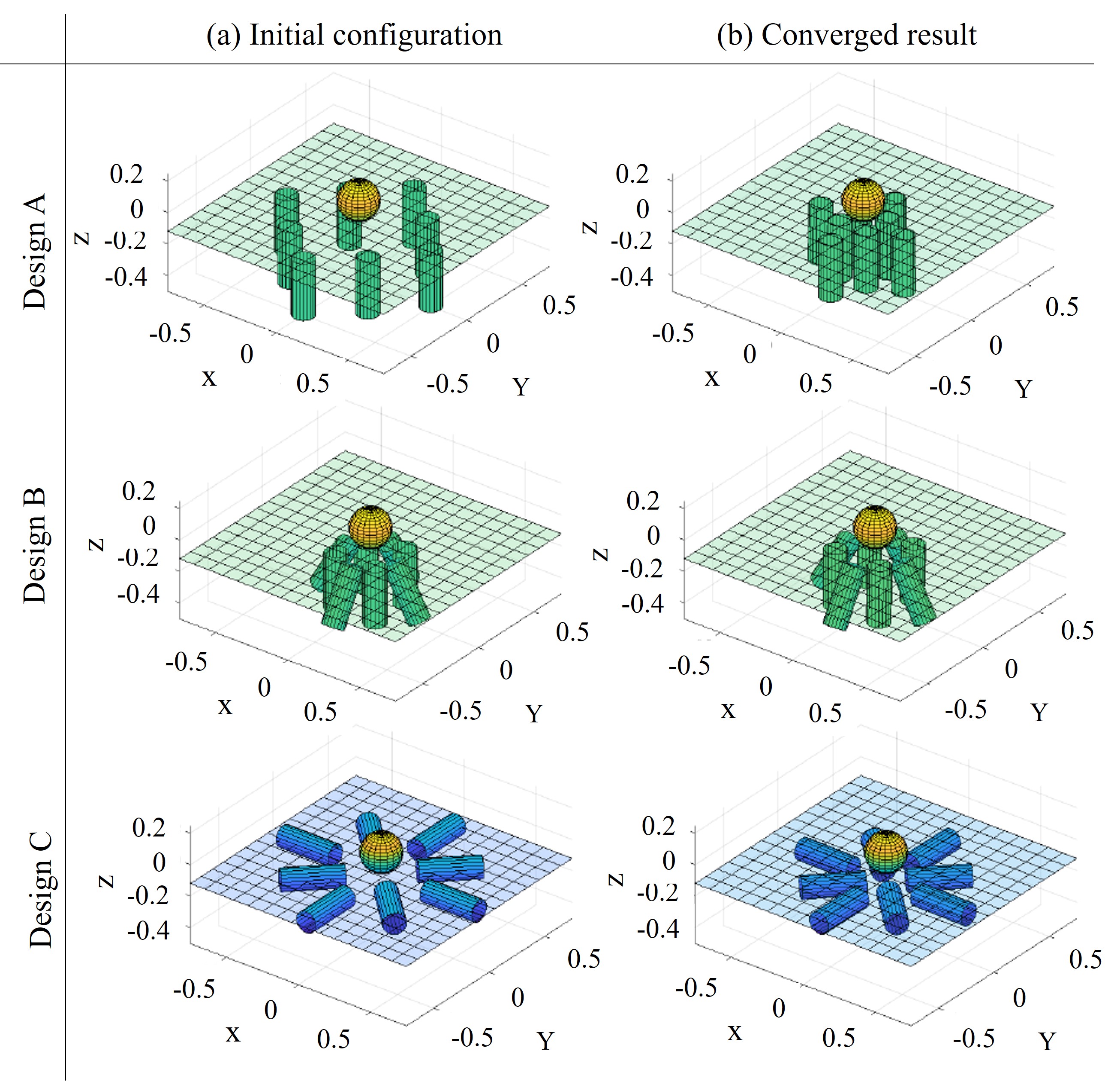}
  \caption{Results of the optimization algorithm, showing the intial configurations in (a) and the associated converged configuration in (b). Top row: the coils are placed in parallel. Middle row: half of the coils are tilted, half of the coils are placed in parallel. Bottom row: the coils are placed in a horizontal plane.}
  \label{Fig-Layout}
\end{figure}

A number of different initial configurations for the electromagnets were chosen to perform the optimization, each of which results in a different local optimum at convergence as shown in Figure \ref{Fig-Layout}.

The predetermined starting configuration for Design A consists of all actuators pointing in the vertical direction and spaced out around the operating space below the surgical table plane. Throughout the optimization simulation, the actuators migrated a fair distance while retaining their symmetry.

Design B starts off in a configuration with half of the actuators at a tilted angle to observe how the algorithm handles actuators that are not pointing along a principle axis. The actuators do move throughout the process, but not by much relative to Design A. 

Design C was initialized with all actuators horizontal and parallel to the operating table and pointing towards the center of the operating space. This initial configuration actually sees the actuators slightly above the plane of the table, and it is clear that the algorithm enforces the planar constraint and lowers the electromagnets to reach the converged results.

In all these designs, the main geometric constraints are satisfied. The general trend seems to be that the actuators all move closer to the operating space throughout the optimization. 

Taking the final system configuration parameters, we compared the performance metrics of optimized Designs A, B, and C. For evaluation we used singular value decomposition to calculate the singular values and condition number (CN) for the control matrix of each design \cite{Abbott-SVD}. Table \ref{Table-Layoutoptimization} reports these singular values and condition numbers, in addition to the fitness metric for each of the three studied designs.  The objective function $\widetilde{F}(\widetilde{\textbf{x}})$ is recorded as a metric for the fitness metric, where a lower number is a better score. The fitness metric is not the best number to use to compare these systems because it also accounts for values such as distance to the work plane and the arbitrary distance between electromagnets. 

\begin{table}[bt]
\centering
\caption{Results of the optimization algorithm for 3 initial configurations}
\begin{tabular}{c|cccc}
\hline
Config. & \begin{tabular}[c]{@{}c@{}}Min Singular \\ Value $\sigma_3$\end{tabular} & \begin{tabular}[c]{@{}c@{}}Min Singular \\ Value $\sigma_1$\end{tabular} & CN   & $\widetilde{\bf F}(\widetilde{x})$ \\ \hline
A & 0.0053 & 0.0087 & 1.64 & 0.033 \\
B & 0.0049 & 0.0054 & 1.10 & 0.045 \\
C & 0.0044 & 0.0068 & 1.55 & 0.054 \\
\hline
\end{tabular}
\label{Table-Layoutoptimization}
\end{table}

Looking at these comparison results, Design B is the most well-conditioned magnetic actuation system layout that was simulated, which implies an ability to create fields in all directions.  Even though Design A is the worst-conditioned configuration, it has the largest minimum and maximum singular values. From these metrics it is reasonable to infer that Design A will produce the largest magnetic fields, even if some directions experience lower fields. In fact, the maximum relative magnetic field was also determined for each design and it was found the value of Design A is $17.2\%$ maximally higher than Design B and $23.3\%$ maximally higher than Design C. Furthermore, when the singular values of the gradient field components were calculated, Design A still has the largest maximum and minimum singular values of all the designs. Addtionally, Design A is the easiest to build given the parallel magnet orientations, and is the most compact design.

Design A was thus chosen for the next prototype electromagnetic actuation system. This system configuration has all electromagnets oriented in the vertical direction with the top faces parallel to the surgical table surface. From a top-down view, the actuators positions are located at the corners of two concentric squares that are misaligned by $45^\circ$ where the smaller inner square has a sidelength of 186 mm and the large outer square has a sidelength of 406 mm. Precise electromagnet center positions are given in Table \ref{Table-EMlocation}.

\begin{table}[bt]
\centering
\caption{Positions of the center of each electromagnet from Configuration A relative to the workspace origin (unit: m)}
\resizebox{\columnwidth}{!}{%
\begin{tabular}{c|rrrrrrrr}
\hline
  & {EM1}  & {EM2}  & {EM3}  & {EM4}  & {EM5}  & {EM6}  & {EM7}  & {EM8} \\ 
\hline
X & -0.203 &  0.000 &  0.203 &  0.131 & -0.131 &  0.203 &  0.000 & -0.203 \\
Y & -0.203 & -0.131 & -0.203 &  0.000 &  0.000 &  0.203 &  0.131 &  0.203 \\
Z & -0.300 & -0.300 & -0.300 & -0.300 & -0.300 & -0.300 & -0.300 & -0.300 \\
\hline
\end{tabular}%
}
\label{Table-EMlocation}
\end{table}

\section{System Implementation}
In this section we report the mechanical, electrical, and software design aspects of the system as-built. The system integration and calibration will also be covered to fully characterize the system. 

\subsection{Mechatronic System}
The mechanical components of the system consist primarily of the table base, electromagnets, table top cover, and cooling system. From the simulation results presented above, we fabricated eight identical electromagnets each with a core diameter of 90 mm, length of 360 mm, and coated with \mbox{22.5 mm} thick copper coil windings. 

The FEA simulations that were conducted above were based on a core material of \textsc{Vacoflux} 50: a cobalt-iron soft ferromagnetic alloy from \textsc{Vacuumschmelze}. In the implementation, we chose the core material as cast iron due to the limited supply and long lead times for such a large quantity of \textsc{Vacoflux} 50. Iron has the same permeability as \textsc{Vacoflux} 50 for low current densities ($< 4$ A$\cdot$mm$^2$). However, \textsc{Vacoflux} 50 maintains constant permeability into higher current densities whereas iron's permeability begins to drop. This effect will be seen in the results presented in the following sections. 

\begin{figure}[t]
  \centering
  \includegraphics[width=0.48\textwidth]{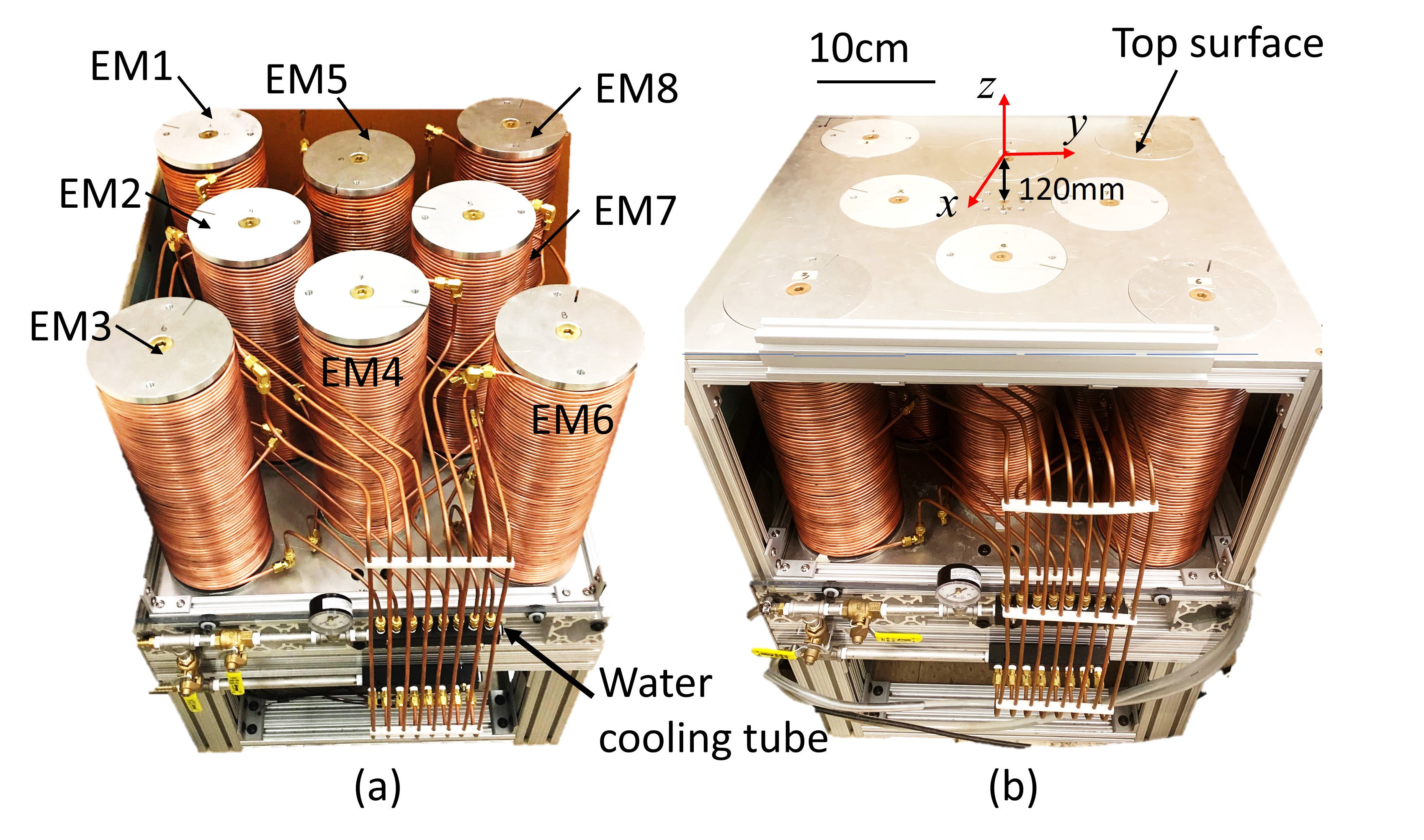}
  \caption{Developed magnetic actuation system: (a) without top surface, (b) with the top surface installed.}
  \label{Fig-CoilModel}
\end{figure}

Since the desired operating current density from the optimization simulation is $J = 6$ A/mm$^2$, an active cooling system to circulate water was required to avoid overheating the system. A soft copper tubing with 3/16$''$ OD and 1/8$''$ ID was wrapped by hand around each electromagnetic actuator and used for circulating the cooling water, as shown in Fig. \ref{Fig-CoilModel}(b). The whole system including the electromagnets and the supporting structure is estimated to weigh 450 kg once assembled. 

Given the power requirements of a single electromagnet being roughly 1.5 kW, a total power of at least 12 kW needs to be supplied to the system to simultaneously activate all actuators at max power. DC power supplies (XP Power) were used to convert AC current to DC current. Eight servo drivers (AB50A100 servo driver - Advanced Motion Controls) were operated in current control mode where a voltage setpoint determines the scaled output current. Closed-loop controllers internal to the motor drivers ensure the requested current is being delivered to each electromagnet. Each electromagnet is equipped with a thermocouple probe placed between the core and the windings which monitors the temperature in the electromagnet during operation. A PCIe Analog and Digital I/O system (Sensoray Model 826) was used to send signals from a desktop computer to the motor drivers and acquire signals from the thermocouples.

\subsection{Magnetic Field Calibration}
To calibrate the control matrix of the system, it is necessary to experimentally determine the relationships between the applied currents and resulting magnetic field components. We chose to calibrate the field only at the system origin (120 mm above the center of the table). 
For each actuator, the current was increased to a maximum positive applied current, reduced back to zero, increased to a maximum negative current, and finally reduced back to zero current while at every point along the way the magnetic field components were measured. This directional sweep of currents should reveal any hysteresis in the magnetic field produced by the iron-core electromagnets. 

Two example results of these sweeps are shown in Supplementary Figure S1. From the calibration results it appears that the core begins saturating towards the higher applied current densities, marked by a drop-off in field linearity. Despite this drop-off, when operating at the max current densities, the actual field strength measured with the gaussmeter is only off by 1-3 mT from the simulated field, which is less than $10\%$ error. 

To calibrate for the gradient components of the magnetic field, a 3-axis motorized gantry system was used to take field measurements at several positions near the center of the system for a given applied current. The probe was moved a small amount $\Delta x$, $\Delta y$ and $\Delta z$ sequentially with new measurements taken at each position. The gradients were then numerically calculated (e.g. for the $x$-component) as
\begin{equation}
    \frac{\partial \bf B}{\partial x} =  \left[
\begin{array}{c}
 \frac{\partial  B_x}{\partial x}  \\
 \frac{\partial  B_y}{\partial y} \\
 \frac{\partial  B_z}{\partial z} 
\end{array}
\right] \approx \frac{ \textbf{B}(x_0+\Delta x) -  \textbf{B}(x_0)}{\Delta x},
\end{equation}
where $x_0 = 0$ is the value of x at the origin.
With all of the relationships between applied currents and the magnetic field magnitude and gradient components, we update the control matrix to control the field based on calibrated field parameters using \eqref{Eq-ControlMatrix-BG}.  The resulting control matrix is given in Supplementary Table I.

The system was further tested with the gaussmeter to determine the maximum magnetic fields that could be produced along each principle axis when using either the uniform control methodology \eqref{Eq-Uni-B} or the non-uniform control methodology \eqref{Eq-nonUni-B}. These system limits will be used to determine the best operating magnetic fields to use to evaluate the magnetic actuation system’s ability to control and actuate a microgripper. These maximum fields are given in Table \ref{Table-Maxfield} for uniform and nonuniform field scenarios, respectively. This table demonstrates the ability to increase the field by a factor of 2-3 times by using a non-uniform field control methodology over the uniform field control methodology.

\begin{table}[bt]
\centering
\caption{Calibrated maximum magnetic fields (Unit: mT)}
\begin{tabular}{cc|ccc}
& & max $B_x$  & max $B_y$ & max $ B_z$ \\ \hline
\multirow{3}{*}{\begin{tabular}[c]{@{}c@{}}Uniform \\ control\end{tabular}} 
& $B_x$ & 11.7 & 0.0 & 0.0 \\
& $ B_y$ & 0.0 & 11.4 & 0.0 \\
& $ B_z$ & 0.0 & 0.0 & 19.3 \\ \hline \hline
\multirow{3}{*}{\begin{tabular}[c]{@{}c@{}}Non-uniform \\ control\end{tabular}} 
& $ B_x$ & 38.0 & 0.0 & 0.0 \\
& $ B_y$ & 0.0 & 38.2 & 0.0 \\
& $ B_z$ & 0.0 & 0.0 & 47.8 \\ \hline
\end{tabular}
\label{Table-Maxfield}
\end{table}

\section{Magnetic Manipulation Experiments}
In this section, we will show the feasibility of the developed coil system to manipulate small magnetic surgical tools.

\subsection{Experimental Setup}
\begin{figure}[t]
  \centering
  \includegraphics[width=0.45\textwidth]{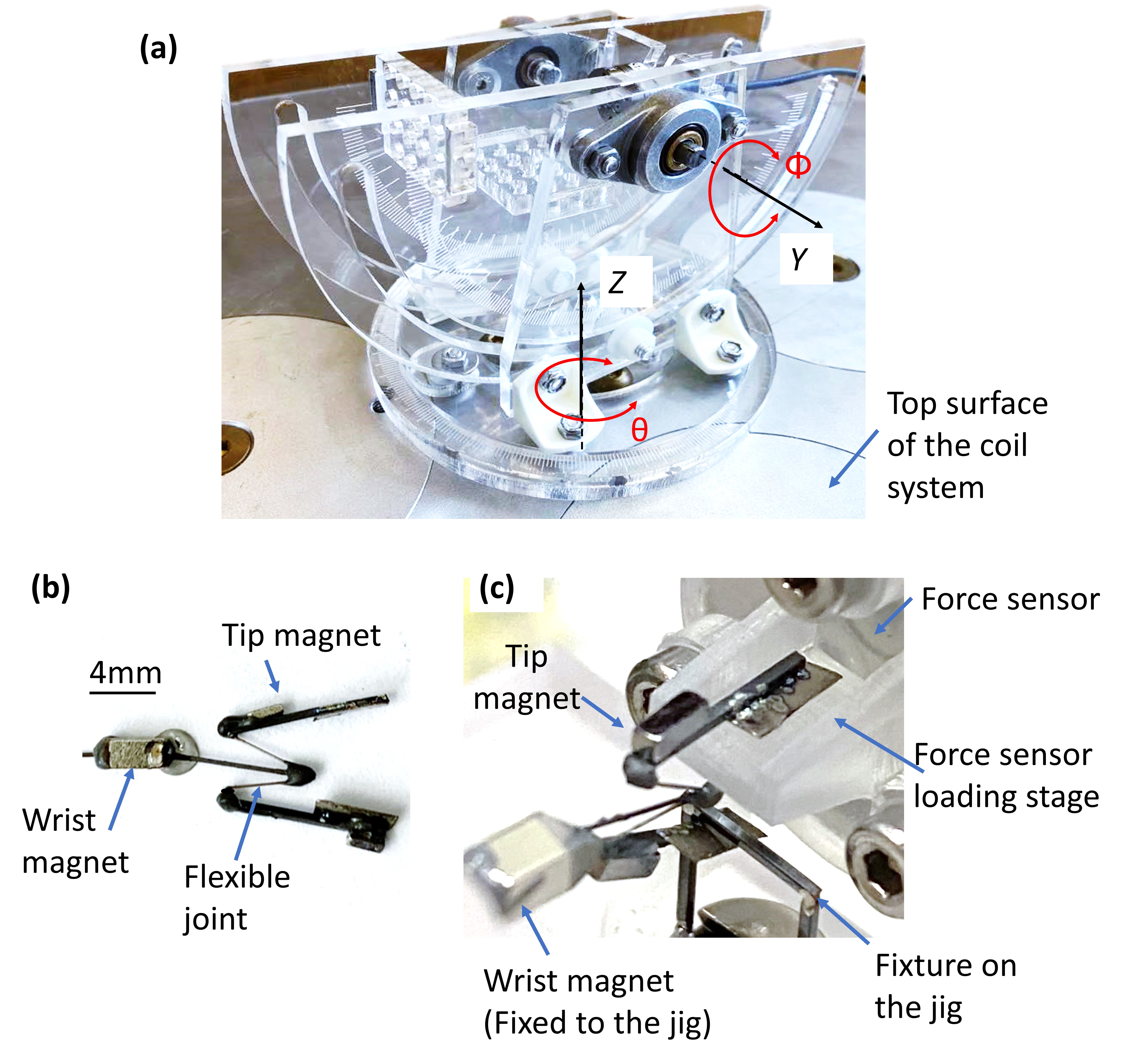}
  \caption{Experimental setup. (a) The custom fixture is made to set the spherical orientation of the microgripper. The polar and azimuthal angles can be set by hand-tightening the bolts on the fixture. This setup allows for $360^\circ$ of rotation about the $z$-axis and up to $ \pm 90^\circ$ of rotation about the $y$-axis using a $ZY$ Euler rotation. (b) The microgripper. (c) The microgripper is mounted on the force sensor.}
  \label{Fig-ExperimentJig}
\end{figure}

Besides the developed coil system, the experimental setup includes a magnetic microgripper developed previously by our group \cite{Forceps-TBME21}, a customized fixture, a 6 DOF force probe (Nano 17 ATI, USA), and laser cut support structures, as shown in Fig. \ref{Fig-ExperimentJig}. The magnetic microgripper has three DOF including gripper grasping and wrist bending in two directions. An N52 grade magnet measuring 1 mm × 1 mm × 3 mm is glued on each of the two gripper jaws and can achieve the grasping motion when external magnetic fields are applied. One additional small magnet is fixed on the wrist joint and used to bend the flexible wrist subject to the magnetic fields. The custom fixture was designed and fabricated from laser-cut acrylic to allow for control over the orientation of the force probe and magnetic microgripper. This fixture allowed the force probe and magnetic device to remain stationary relative to each other, while allowing both to be rotated about the $z$- and $y$-axes. 
The wrist of the microgripper is clamped down to the fixture and the open ``jaws" of the gripper grasp around the force probe and a stationary restraint. The restraint is necessary for the gripper to pull against or when a field is applied, the gripper will grab the probe and the equal but opposite grasping forces will result in force reading. When the gripper is in this setup it needs to be opened slightly to fit around the jaws around the probe. This pre-load is offset in the measured force values. 

\subsection{Demonstrations}
Three experiments were performed to demonstrate the system’s ability to actuate the microgripper to apply a large grasping force under different conditions.

\subsubsection{Experiment I: Gripper placed in horizontal orientation}
The purpose of this experiment is to determine the actuation strength of the tool in the horizontal orientation with an increasing applied magnetic field. The horizontal orientation chosen had the orientation angles set to $\theta = 0^\circ$ and $\phi= 0^\circ$ which corresponds to the microgripper magnets’ long axes pointing horizontally parallel to the $x$-axis, and the magnets’ magnetization axis pointing upwards in the $z$-direction.

Before the experiment, ten datapoints were recorded while the system was not applying any magnetic field to the device in the workspace. The average of these values is used as a relative offset. The experiment was conducted once with the use of the non-uniform field control methodology using \eqref{Eq-nonUni-B} and repeated using the uniform field control methodology via \eqref{Eq-Uni-B} to compare the performance of the tool when being actuated differently. For the uniform field control methodology, a uniform magnetic field in the positive x-direction $\textbf{B}_x$ was applied which increased by 1 mT at every datapoint to a max of 10 mT. For the non-uniform field control methodology a non-uniform magnetic field $\textbf{B}_x$ was applied at intervals of 5 mT from 0 mT to a max of 30 mT. For each magnetic field applied, ten datapoints of the force and torque components in each cartesian direction were recorded. 

\begin{figure}[t]
  \centering
  \includegraphics[width=0.48\textwidth]{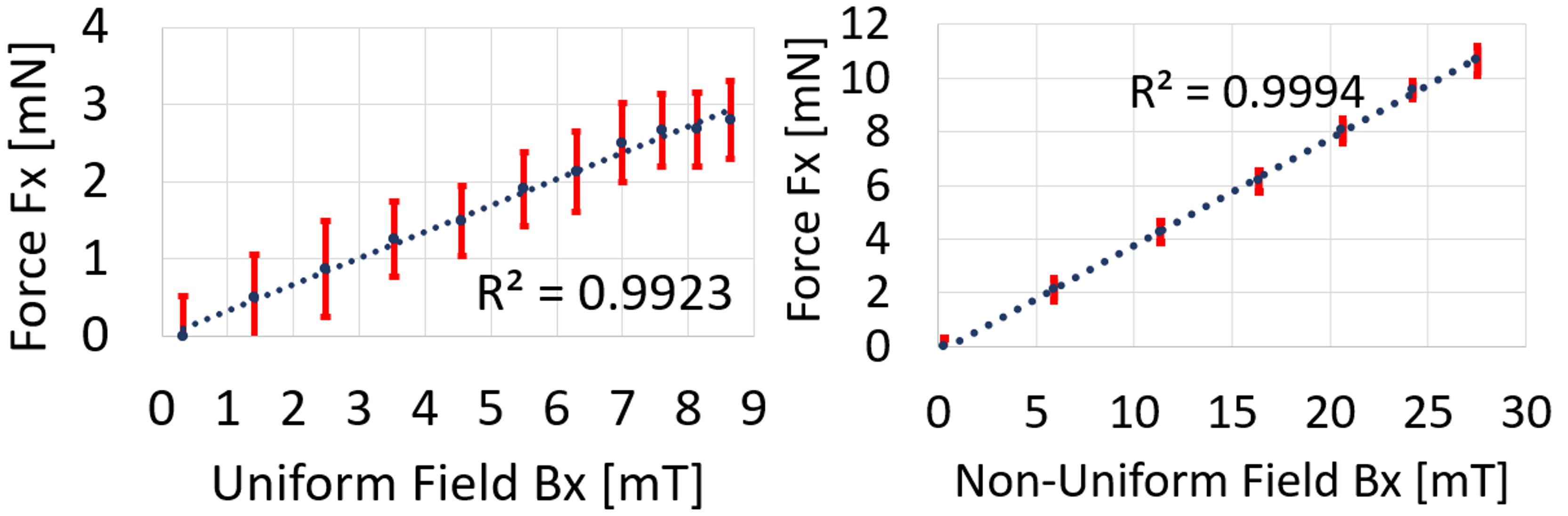}
  \caption{Results of grasping force vs applied magnetic field for a \textbf{horizontally oriented} microgripper. Actuation strengths are reported by the slope relating magnetic field to grasping force. These plots show a small increase in actuation strength when using non-uniform fields.}
  \label{Fig-Results-1}
\end{figure}

\subsubsection{Experiment II: Gripper placed in vertical orientation}
This experiment aims to determine the actuation strength of the microgripper in the vertical orientation to compare the effects of using a non-uniform magnetic field instead of a uniform magnetic field. To orient the magnets in the vertical direction, only the y-axis carriage need to be rotated such that the rotation angles corresponded to $\theta$ = 0 deg and $\phi$ = 90 deg. In this orientation, the magnets’ magnetization vector is pointing along the $x$-axis and their long axes are pointing vertically, along the negative $z$-axis.

Similar to the horizontal experiments, ten measurements were recorded with zero applied field to establish a relative offset, and two control modes were tested. For the uniform control methodology experiment, a uniform field $\bf B_z$ was applied in the negative z-direction at 1 mT interval to a max of 15 mT while ten datapoints of force and torque readings were recorded at each interval. For the non-uniform field control methodology experiments, a non-uniform field $\bf B_z$ was applied in the negative $z$-direction at intervals of 5 mT from 0 mT to a max of 40 mT with five datapoints being recorded for each different applied field.

\subsubsection{Experiment III: Pick-and-place demonstration}
The final experiments investigated the grasping effect of the microgripper. The microgripper was fixed at the base and its tip was located around the origin of the electromagnetic table. A small piece of Styrofoam was placed below the gripper. In the experiment, the microgripper was controlled to point down to grasp and lift the cargo, and then place the cargo back on the holder. The experiment process was recorded for qualitative evaluation.

\begin{figure}[t]
  \centering
  \includegraphics[width=0.48\textwidth]{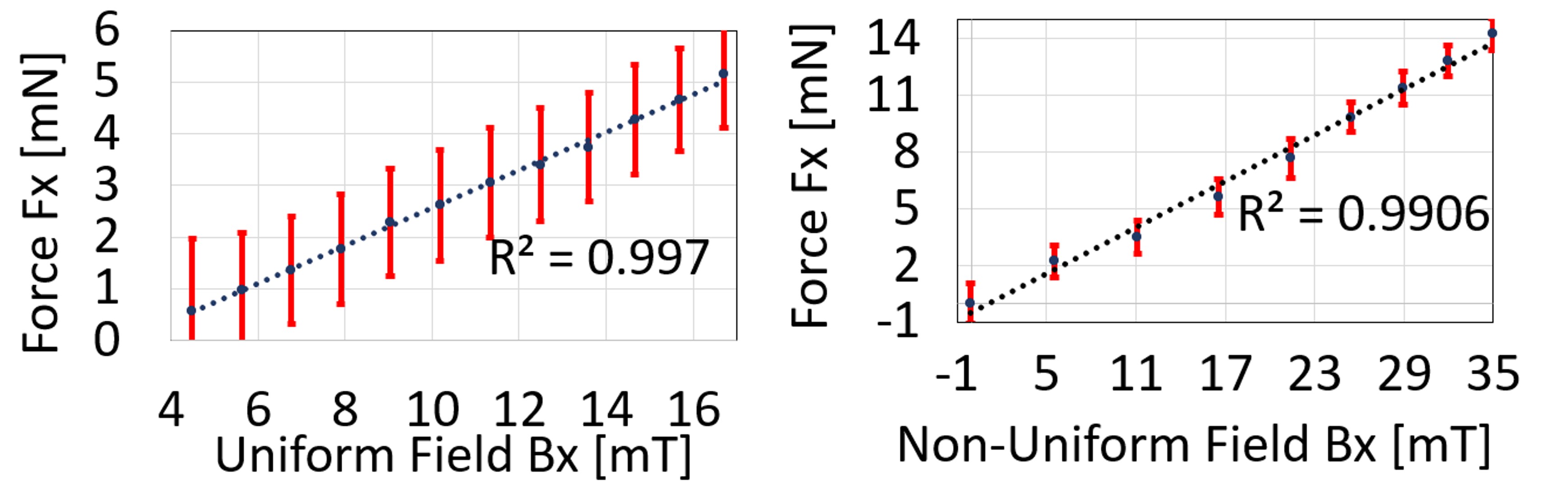}
  \caption{Results of grasping force vs applied magnetic field for a \textbf{vertically oriented} microgripper. Actuation strengths are reported by the slope relating magnetic field to grasping force. These plots show a small increase in actuation strength when using non-uniform fields. The over maximum grasping force recorded was $14.3 \pm 0.8$ mN for non-uniform magnetic fields.}
  \label{Fig-Results-2}
\end{figure}

\subsection{Results}
\subsubsection{Experiment I: Gripper placed in horizontal orientation}
Figure \ref{Fig-Results-1} shows the measured relationship between an applied parallel magnetic field and the resulting force the gripper applies to the force-torque probe when the device is orientated horizontally. 
The force results reported are the combination of two force-torque readings. Since the force-torque sensor measures the three force vector components and the three torque vector components, the torque about the $y$-axis was factored into the force in the $x$-direction using the following formula:

\begin{equation}
     F_{\text{grip}} =  F_x + \frac{\uptau_y}{r_{\text{grip}}}
\end{equation}
where $r_{grip} = 10$ mm is the estimated finger beam length. In these results, the addition of the torque component had a small effect on the initial $F_x $ data, increasing the force measurements by roughly  $15\%$.

A linear trend was fit to both plots for uniform applied fields and for non-uniform applied fields to quantify the actuation strength as a force per applied magnetic flux density. For the applied uniform fields, the actuation strength was \mbox{0.340 ±0.010 N/T} which is comparable to the actuation
strength of 0.309 N/T determined by Lim et al in \cite{Forceps-TBME21} for the same designed device. Furthermore, we can see that the actuation strength achieved using a non-uniform control methodology can increase the actuation strength to \mbox{$0.397 \pm0.004$ N/T}. The maximum achievable grasping force for the uniform methodology was only $2.8 \pm 0.5$ mN since the max field only reached between 8-10 mT. In comparison, the maximum grasping force measured in \cite{Forceps-TBME21} was 6.1 mN or roughly twice as large. However, when the
system is controlled via a non-uniform control methodology, a maximum of $10.6 \pm 0.5$ mN was measured which is a sizable improvement over previous actuation forces.

\subsubsection{Experiment II: Gripper placed in vertical orientation}
The measured relationship between an applied parallel magnetic field and the resulting grasping force when the device is oriented vertically is shown in Figure \ref{Fig-Results-2}. The same calculations as before were performed to combine the torque and force components into a single grasping force measurement.

In this dataset, the first three data points from the uniform field control methodology were omitted because the probe did not read any force or torques. In this vertical orientation case, the gripper was not initially in contact with the probe and needed to be actuated by the magnetic field to grasp the probe before readings occurred. These datasets were fit with a linear trendline to determine the gripper’s actuation strength in these two different control scenarios. The uniform fields resulted in an actuation strength of $0.367 \pm 0.006$ N/T for the vertically oriented magnetic device which is roughly the same as the horizontal configuration for the same control method. The non-uniform field control resulted in an actuation strength of $0.408 \pm 0.015$ N/T which is again comparable to the actuation strength found for the gripper in the horizontal configuration for the same control methodology. Since the system is better at generating a large magnetic field in the vertical direction compared to the horizontal directions, the system was able to apply a large magnetic field in these vertical experiments. This results in a maximum grasping force of $14.3 \pm 0.8$ mN being measured which is the largest controllable force this design has seen in an actuation system.

The results of the vertical and horizontal actuation strength experiments for uniform and non-uniform control methodologies are summarized in Table \ref{Table-GripperForce}.

\begin{table}[bp]
\caption{Summary of the actuation strength of the microgrippers for different orientations and applied field control.}
\centering
\begin{tabular}{l|l|l}
\hline
                       & Non-uniform Fields & Uniform Fields     \\ \hline
Vertical Orient.   & $0.40\pm0.015$ mN/mT & $0.36\pm0.006$ mN/mT \\ \hline
Horizontal Orient. & $0.39\pm0.004$ mN/mT & $0.34\pm0.010$ mN/mT \\ \hline
\end{tabular}
\label{Table-GripperForce}
\end{table}

\subsubsection{Experiment III: Pick-and-Place Demonstrations}

\begin{figure*}[t]
  \centering
  \includegraphics[width=0.95\textwidth]{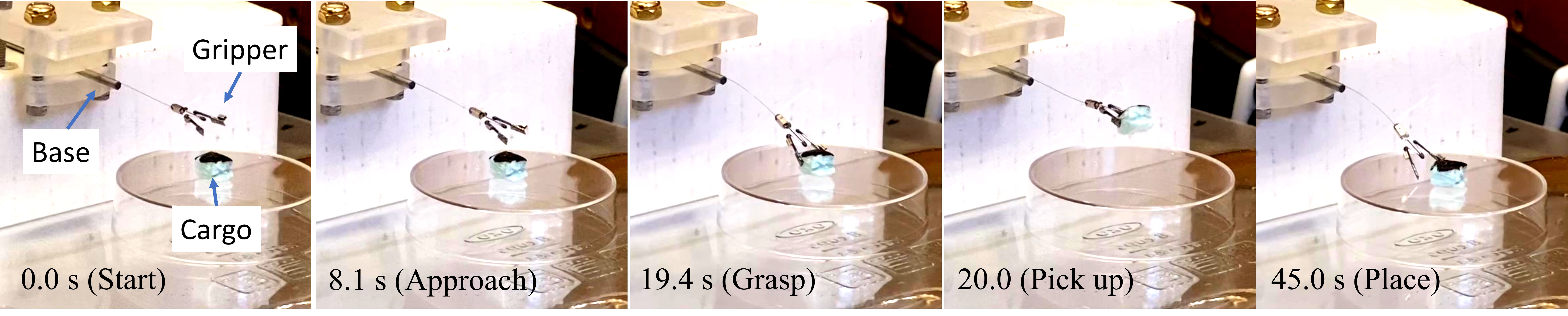}
  \caption{Frames from a video demonstrating pick-and-place operation with the microgripper. The distance from the gripper in the first frame to the table surface is 110 mm. Timestamps are included on the frames, and the entire procedure took less than 45 s.}
  \label{Fig-PickPlace}
\end{figure*}

This demonstration showcases the openness of the tool control where the gripper can be seen from almost all perspectives. As shown in the \textit{supplementary video}, the tool was actuated to turn left and right (yaw) as well as in the up and down (pitch) motions to demonstrate control over the wrist motions. The gripper was able to lift it, move it side to side and upwards before placing it back down. Frames from the recorded video are shown in Figure \ref{Fig-PickPlace} with timestamps. The whole demonstration takes place within 45 s, which is an acceptable actuation speed for this device.

\section{Conclusion and Discussion}

In this paper, we designed and built a large-scale magnetic actuation system for controlling magnetic microrobots in a mock surgical environment. We investigated the relationship between magnetic actuator mass and the maximum magnetic field that it can generate, and presented a design methodology using a gradient descent-based approach to optimize several constraints and parameters. We determined that the workspace accessibility is $222^\circ$ using the solid angle formulation we defined in section \ref{Section-WorkspaceAccessibility}. This accessibility is greater than the other stationary 8-electromagnet actuation systems found in the literature, and qualitatively the operating space is less obstructed than other systems as well. Using a gaussmeter, all the control variables for the magnetic field and the magnetic field gradient components were calibrated to fully characterize the system and its performance. Post-calibration, the system was able to generate non-uniform magnetic fields up to 38 mT in the $x$- and $y$- axes and 47 mT in the $z$-axis at a working distance of 120 mm. This complete system was then used to investigate its ability to control the microgripper under different field control methodologies. 

The experimental results highlight the advantages of using a non-uniform field control methodology to increase the maximum grasping force of the microgrippers, for electromagnetic navigation systems with a high degree of coil constraints. We also show the potential increase in magnetic field magnitude that can be achieved by using a non-uniform control methodology over the conventional uniform field control methodology. In the experiments, we selected a tethered microgripper for tissue grasping as the agent to demonstrate our table-like coil system's actuation capability. However, the applications can be extended to actuate and navigate a broad range of small-size (millimeter scale) magnetic tools, for instance, magnetic catheters and untethered cargo robots. The actuation of the gripper is demonstrated in two representative orientations (horizontal and vertical). However, the gripper can be controlled in any pose by applying the field in the proper direction according to (\ref{Eq-MagTorque}).

Returning to our motivation for workspace accessibility, if we only consider the actuators and not the support structure as we did for the other systems, a solid angle of $222^\circ$ is measured from the perspective of the microrobot in the center of the workspace. This metric is bounded by the outermost electromagnets and will improve if we were to arbitrarily set the workspace further away from the actuation system. This workspace accessibility metric is only $2^\circ$  larger than the estimate we determined for the MiniMag system \cite{minimag} but the operating space itself is significantly larger. Unlike the actuation systems in the literature, it is clear from photographs of this system that an object can be placed in the workspace center from most angles above the table surface due to the actuators not being concave around the workspace. Furthermore, if the support structure is included, this performance metric decreases even more to $208^\circ$. Regardless, since a patient can freely fit in the operating space, this system has met its geometric constraint goals.

In an ideal optimization simulation, the initial values for the optimization parameters are not critical since there would be only one global minimum that the function would reach. With our simulation, this is far from the case. This simulation converges to many different local minima depending on the initial starting parameters, which could possibly be due to the large number of DOF and optimization parameters. The fact that we have more DOF than we do constraining equations may also contribute to the multiple best solutions. To find the best minimum solution, many different trials were conducted with the results recorded. For the majority of simulations, a random input would result in an equally random looking output, which would be very difficult to fabricate and whose objective function did not have a low final cost. When predetermined starting configurations were set, the resulting configuration also looked uniform and symmetric.

\bibliographystyle{IEEEtran} 
\bibliography{main} 




\end{document}


\title{Supplementary}

\maketitle

\IEEEpeerreviewmaketitle

\section{Optimization functions}

\subsection{Magnetic Field Optimization Functions}
The magnetic field optimization functions $\widetilde{\bf M}(\widetilde{\bf x})$ are responsible for attempting to maximize the magnetic fields produced by the system. From Equation 7 we can calculates the currents required to produce some arbitrary desired magnetic field in one of the principle axes. Just as with the first attempt at this associated function, the magnetic control matrix $\mathcal{B}({\bf P})$ is required to be normalized by the input currents, such that a current 1 corresponds to the max operating current. The maximum magnetic field is determined by scaling the initial arbitrary principle magnetic field by the maximum current required so that all the currents $|I| \leq 1$:  $\bf B_{\text{max}} = \bf B_{des}/\text{max}(|\bf I|)$. The desired field is arbitrary and set to unity while the currents I are substituted out: $\bf B_{\text{max}}=1/\text{max}(|\mathcal{B}({\bf p})^{\dag} \bf B_{des}|)$. This equation gives the maximum nonuniform field that can be produced in any general specified direction.

For the associated function, directions of importance are the 3 principle axes given by their cartesian unit vectors. Furthermore, to maximize the magnetic field using gradient descent, we will instead minimize the inverse of the max field. These resulting functions are shown below where the max () function returns the maximum value of the resulting vector.

\begin{equation}
    \widetilde{\textbf{M}}(\widetilde{\textbf{x}}) = \left[
\begin{array}{c}
\epsilon_xB_{max,x}^{-1} \\
\epsilon_yB_{max,y}^{-1}\\
\epsilon_zB_{max,z}^{-1}
\end{array}
\right] = \left[
\begin{array}{c}
\epsilon_x \text{max}(|\mathcal{B}(\bf p)^{\dag} B_{des,x}|) \\
\epsilon_y \text{max}(|\mathcal{B}(\bf p)^{\dag} B_{des,y}|) \\
\epsilon_z \text{max}(|\mathcal{B}(\bf p)^{\dag} B_{des,z}|) 
\end{array}
\right]
\end{equation}

The optimization functions are all scaled by a factor $\epsilon_i$ corresponding to 0.002 to aid in the optimization convergence. These factors and others in the following associated functions, are arbitrary and were determined through some trial and error while observing the relative weighting of each function so that one associated function does not completely dominate the rest. We used the scaled dipole moment function for each actuator to calculate these metrics for the maximum magnetic field in this optimization routine. 

\subsection{Electromagnet Height Penalty Functions}
The electromagnet height penalty functions $\widetilde{\bf H}(\widetilde{\bf x})$, attempt to enforce the height constraint by penalizing the cost function. This function returns a minimum when the electromagnet is in contact with the upper bounding plane with the expectation that the resulting configuration will consist of actuators as close to the patient as possible.
The position of each electromagnet is defined by a point at its centroid. To determine if the electromagnet is touching the top of the surgical table (from below) a relationship between the height $h^\prime$ and the electromagnet’s $z$ position and orientation was derived. Figure 1 labels the important actuator parameters including the diameter d and the length $l$.

The height relation that was derived is shown below.
\begin{equation}
\begin{split}
    h(z_i, \gamma_i) &=  z_i +h^\prime \\&=  z_i+\frac{1}{2}\sqrt{l^2+d^2}\text{sin}(\frac{\pi}{2}-(\gamma_i-\text(tan)^{-1}(\frac{d}{l}))),
\end{split}
\end{equation}
where $z_i$ is the centroid position along the Z axis of the actuator.

With this general form of the associated function we can construct the full set of electromagnet height penalty functions as:

\begin{equation}
    \widetilde{\bf H}(\widetilde{\bf x}) = \left[
\begin{array}{c}
-\text{log}(-h(z_1,\gamma_1)/z_0) \\
-\text{log}(-h(z_2,\gamma_2)/z_0)\\
\vdots \\
-\text{log}(-h(z_n,\gamma_n)/z_0)
\end{array}
\right],
\end{equation}
where $z_0$ is the height of the bounding plane relative to the origin equal to 120 mm. Since the height value is non-zero when the actuator is touching the surface of the bounding plane, a barrier function using -log was implemented to severely penalize the function if the actuator’s position is too high.

\subsection{Electromagnet Proximity Penalty Functions}
Without some constraint or penalty on the positions of the electromagnets relative to one another, all actuators will amalgamate, occupying the same space, to all be closer to the workspace and increase the magnetic fields. The proximity penalty functions $\widetilde{\bf P}(\widetilde{\bf x})$ aim to determine the minimum distance between each actuator and its neighbours and push them away from each other to avoid collisions.

To determine the distance between two actuators modeled as cylinders, we will first assume the actuators are infinitely long and find the minimum separation between the two extended lines. Checks will then be made to determine if the minimum difference is the true separation distance between the two finite cylinders.

Any two electromagnets can be modelled as two finite lines with a starting point \textbf{s} and extending in the direction of $\hat{\textbf m}$ parameterized by
\begin{equation}
    \textbf{p}_i=\textbf{s}_i+||\textbf {m}||\hat{\textbf m}_i = \textbf{s}_i+\textbf{m}_i.
    \label{Eq-PositionElectromaget}
\end{equation}
The distance between two electromagnets can be calculated directly with Equation \ref{Eq-PositionElectromaget} when the two actuators are parallel. In the case that the extension lines of the two actuators are skewed, The distance between the two skewed lines is the projection of the unit vector perpendicular to both lines onto the vector between the two lines described by the following equation:
\begin{equation}
    D_{ij} = 	\left| \frac{\textbf{m}_i \times \textbf{m}_j}{||\textbf{m}_i \times \textbf{m}_j||} (\textbf{s}_j - \textbf{s}_i)  \right|
\end{equation}

From this relation three outcome scenarios are possible: the minimum distance vector will intersect both finite cylinders, intersect only one of the cylinders, or intersect neither cylinders. To determine if this minimum distance is on the finite section of the electromagnet cylinder, we will determine the points of intersection and determine if the point is within the length of the cylinder. Using $\pmb {\lambda}=\textbf{m}_i \times \textbf{ m}_j$, to define the expressions $\pmb {\lambda}_i=\textbf {m}_i \times \pmb {\lambda}$ and $\pmb {\lambda}_j=\textbf{m}_j \times \pmb {\lambda}$, the points of intersection $\textbf {C}_i$ and  $\textbf {C}_j$ for each electromanget can be calculated using 
\begin{equation}
    \textbf {C}_i = \textbf {s}_i + \frac{(\textbf {s}_j-\textbf {s}_i)\pmb {\lambda}_j}{\textbf{m}_i \pmb {\lambda}_j}\textbf{m}_i.
\end{equation}
These two points of intersection can be used to check if they fall on the cylinder axis. If the minimum separation vector does intersect with actuator $i$ then $||\textbf{C}_i-\textbf{s}_i||\leq \frac{l}{2}$ will be satisfied and if the vector intersects with actuator \textit{j} then $||\textbf{C}_j-\textbf{s}_j||\leq \frac{l}{2}$. If both are satisfied, $D_{ij}$ is returned as the minimum separation distance. If none of the intersection points lie on either cylinder, then the minimum distance from the ends of the finite cylinders is calculated by returning the minimum of all combinations of $||[\textbf{s}_i \pm \frac{l}{2}\hat{\textbf {m}}_i ] - [\textbf{s}_j \pm \frac{l}{2}\hat{\textbf {m}}_j ]||$. If just one intersection point does not lie on one of the cylinders, the minimum of the two combinations of $||\textbf{C}_i - [\textbf{s}_j \pm \frac{l}{2}\hat{ \textbf{m}}_j ]||$ are returned. 

With all cases covered a value for the minimum axis-to-axis separation distance $D_{ij}$ will be returned. To get the minimum distance between the surfaces of the cylinders a value of two radii \textit{d} is subtracted from the total separation distance. This is an approximation for cases were one or more intersection points do not fall within their respective actuator, but it should be a conservative estimate for at least half of cases. To maximize the space between electromagnets, the inverse of this difference is minimized. The electromagnet proximity penalty functions are presented as

\begin{equation}
     \widetilde{\textbf{P}}(\widetilde{\textbf{x}}) = [\sigma (D_{ij}-d)^{-1} ]
\end{equation}
where ${i,j}$ has elements in the set of all 2-combinations of the set $S$ where the set $S\in \{1,2,...,n \}$ representing all combinations of distances between electromagnets with no repeats. The scaling factor of $\sigma$ was arbitrarily set to 0.001 to aid in convergence and have the same magnitude as the other associated functions.

\section{Calibration results}
The resulting control matrix is given in Table \ref{tab:ctrlMat} and two example results of calibrated current-field correlations are shown in Figure S1. 

\begin{table*}[!b]
\centering
\caption{The resulting control matrix in units of mT/A and mT/m$\cdot$A.}\label{tab:ctrlMat}
    \begin{equation*}
        \begin{bmatrix}
            \mathbf{B}(\mathbf{0})\\
            \mathbf{G}(\mathbf{0})
        \end{bmatrix} = 
        \mathcal{U}(\mathbf{0})\mathbf{I} = \begin{bmatrix*}[r]
             0.15&  -0.03& -0.17& -0.72&  0.73& -0.15&  0.07&  0.17\\
             0.16&   0.76&  0.15& -0.04&  0.03& -0.17& -0.71& -0.15\\
            -0.03&   0.52& -0.05&  0.51&  0.51& -0.05&  0.51& -0.05\\
            -0.64&   6.40& -0.80& -3.31& -3.90& -0.51&  6.45& -0.96\\
            -1.60&   0.15&  1.72& -0.26& -0.04& -1.55&  0.66&  1.52\\
            -0.35&   0.64&  0.42&  9.64& -9.47&  0.29& -0.48& -0.46\\
            -0.78&  -3.75& -0.60&  6.22&  6.83& -0.85& -4.02& -0.58\\
            -0.45& -10.32& -0.39&  0.38& -0.86&  0.41&  9.60&  0.37
        \end{bmatrix*}\mathbf{I}
    \end{equation*}
\end{table*}

\begin{figure*}[!tb]
  \renewcommand*{\thefigure}{S\arabic{figure}}
  \centering
  \includegraphics[width=0.95\textwidth]{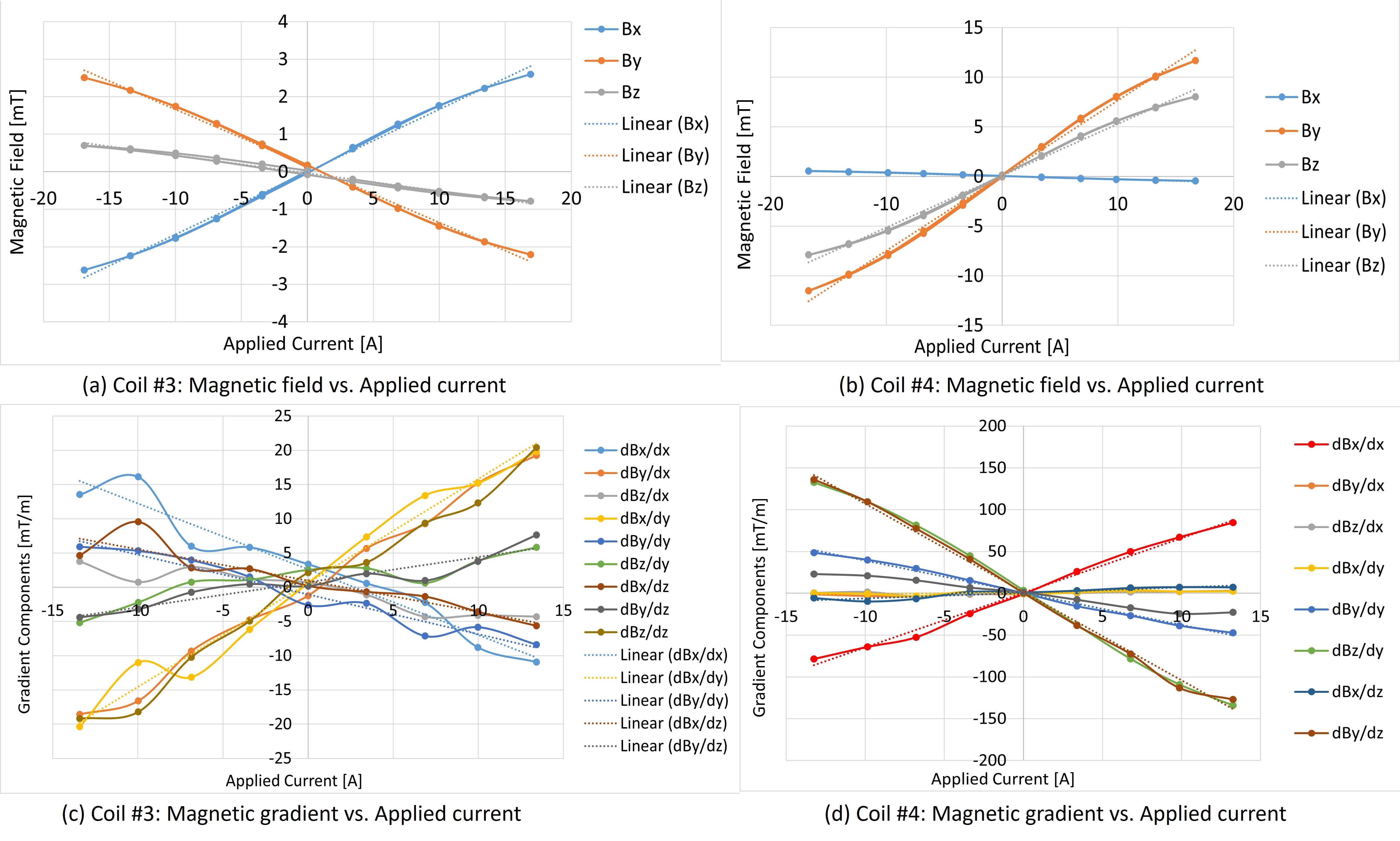}
  \caption{Calibration experiments for characterizing the magnetic field generation (Top row) and gradient generation (Bottom row) of two electromagnets. For each input current the magnetic fields were measured at the origin using a gaussmeter, then the probe was moved a small amount to numerically calculate the field gradients.}
  \label{Fig-CalibrationCurve}

\end{figure*}

